\documentclass[10pt,twocolumn,letterpaper]{article}

\usepackage{iccv}
\usepackage{times}
\usepackage{epsfig}
\usepackage{graphicx}
\usepackage{amsmath}
\usepackage{amssymb}
\usepackage{multirow}
\usepackage[accsupp]{axessibility}  

\usepackage[breaklinks=true,bookmarks=false]{hyperref}

\iccvfinalcopy 

\def\iccvPaperID{12} 

\ificcvfinal\fi

\begin{document}

\title{Balanced Masked and Standard Face Recognition}

\author{Delong Qi, Kangli Hu, Weijun Tan*, Qi Yao, Jingfeng Liu\\
Shenzhen Deepcam Information Technologies, Shenzhen, China\\
{\tt\small \{delong.qi,kangli.hu,weijun.tan,qi.yao,jingfeng.liu\}@deepcam.com}\\
*LinkSprite Technologies, USA\\ \tt\small weijun.tan@linksprite.com
}

\maketitle
\ificcvfinal\thispagestyle{empty}\fi

\begin{abstract}
   We present the improved network architecture, data augmentation, and training strategies for the Webface track and Insightface/Glint360K track of the masked face recognition challenge of ICCV2021. One of the key goals is to have a balanced performance of masked and standard face recognition. In order to prevent the overfitting for the masked face recognition, we control the total number of masked faces by not more than 10\% of the total face recognition in the training dataset. We propose a few key changes to the face recognition network including a new stem unit, drop block, face detection and alignment using YOLO5Face, feature concatenation, a cycle cosine learning rate etc. With this strategy, we achieve good and balanced performance for both masked and standard face recognition.  
\end{abstract}

\section{Introduction}

Face recognition, as a method to identify or verify a person's identity using a face image, has made tremendous progresses in recent years, thanks to deep learning, particularly deep CNN. However, there are still some challenges, one of which is low quality of images. The factors that cause low quality images include pose, blur, occlusion, illumination etc. 
As the outbreak of the disastrous COVID-19 and inspired by the motivation to prevent virus spreading, face recognition has become one way to trace COVID-19 patients and close contacts \cite{COVID-trace}. After a patient is confirmed and his identity is recognized, safety measures can be taken to control the virus spreading. Face mask, as an effective way to prevent the virus spreading, poses a new challenge to traditional face recognition. Since the mask covers a large part of face where abundant features are present, traditional face recognition algorithm may not work effectively. This drives a need to understand how face recognition algorithm deals with masked faces, as a special case of occlusion.  

To cope with the challenge arising from wearing masks, it is crucial to improve the existing face recognition algorithms. Even though some commercial providers have claimed the availability of face recognition algorithms capable of handling face masks, and an increasing number of research publications have surfaced, there is still no publicly available masked face recognition benchmark. The ICCV2021-MFR Workshop organises Masked Face Recognition (MFR) challenge \cite{deng2021mfrinsightface, MFR, zhu2021mfrwebface, webface260m} and the goal is to bench-marking face recognition algorithms for masked faces. 

In this paper we present our solution for this MFR challenge \cite{MFR}. We notice that one challenge is how to balance the performance of the masked and standard face recognition. At the beginning of the challenge, the MFR error rate was used as the ranking metric. To achieve best MFR performance, people tend to put a lot of masked face images in their training dataset. This causes a overfitting problem such that the performance of the MFR is very good, but that of the standard face recognition is getting worse. Later the organizer realized this problem and changed there ranking metric to a mixture of the error rates of the MFR and standard face recognition. In our solution, we realize this problem from the beginning and keep the balance of the MFR and standard face recognition. Our strategy is to control the percentage of the masked face image, by no more than 10\% of the total number of face images in the training dataset.  

Other than this key strategy, we propose a few changes to the backbone network architecture. We use the ArcFace \cite{ArcFace} as our baseline with Resnet \cite{Resnet} as the backbone. We test the ArcFace loss \cite{ArcFace}, cosFace loss \cite{CosFace}, and other loss functions. We design a new stem unit, and add the drop block \cite{dropblock} in the last two layers of the Resnet network. We also propose to use YOLO5Face \cite{YOLO5Face} to do face alignment, and to concatenate features from multiple models to improve the recognition accuracy.   

As of August 3 2021, we achieve the All-Masked MFR metric 0.1056 in the Webface track and TAR@Mask 0.84327, TAR@Mask-All 0.92702 in the Insightface/Glint360k track.  

\section{Related Work}

Face recognition has been widely studied and deployed in practical applications in recent years. For a latest review, please refer to \cite{survey}. For face recognition in the wild, more and more attentions are on poor quality face images by pose, blur, occlusion, illumination etc. Some examples are \cite{NAN,universal,eqface}.  

MFR is a challenging task since a large part of the face is occluded, so abundant features on mouth, nose, and lower cheek are all lost. As a result, the face recognition algorithm has to focus on the eyes, ears, upper cheek, forehead to identity a person's identity.     

The work \cite{anwar2020masked, FMA-3D} provide a tool to generate masked face images as a data augmentation for MFR.  

In \cite{singlecameraMFR}, normal face detector is used to detect masked face, and a pretrained VGGFace2 \cite{vggface2} is used to extract features for face recognition. They treat the MFR as a normal face recognition problem.  

In \cite{hariri2021efficient}, the author first removes the masked region, then use pre-trained networks to extract the features of the eyes and forehead, and finally use a bag of words and a MLP to do the classification. 

In \cite{montero2021boosting} a Resnet50 \cite{Resnet} and ArcFace \cite{ArcFace} are used. They introduce an probability of mask usage, and add a mask usage classification loss to the ArcFace loss.  

In \cite{LPD}, the authors collect some MFR dataset, and propose a latent part detection to locate the latent facial part which is robust to mask wearing, and the latent part is further used to extract discriminative features. 

In \cite{cropping_attention} the authors explore the Convolutional Block Attention Module in a Resnet50 network. They also suggest removing the masked region helps the MFR. They test their algorithm on a variety of MFR datasets.

\section{Methods}

All major changes are described in this section. These include the data augmentation for masked face images, a new stem unit, a DropBlock, YOLO5Face \cite{YOLO5Face} detection and alignment, and feature concatenation. 

\subsection{Data Augmentation for Masked Face Images}

Generation of masked face images is one type of data augmentation methods. However, for the importance of a balanced masked and standard face recognition, we put it here as a separate subsection. 
In our early study, we find that in order to get the best performance of MFR, people is temped to use as many as possible masked face images in the training dataset. As a result, the face recognition model is over-fitted for MFR, but does not work well on normal face recognition. 

Therefore, we control the balance of the MFR and standard face recognition by controlling the percentage of masked face images in the training dataset. After some experiment we find that 10\% is a good trade-off.  

\begin{figure}[t]
    \centering
    \includegraphics[scale=0.35]{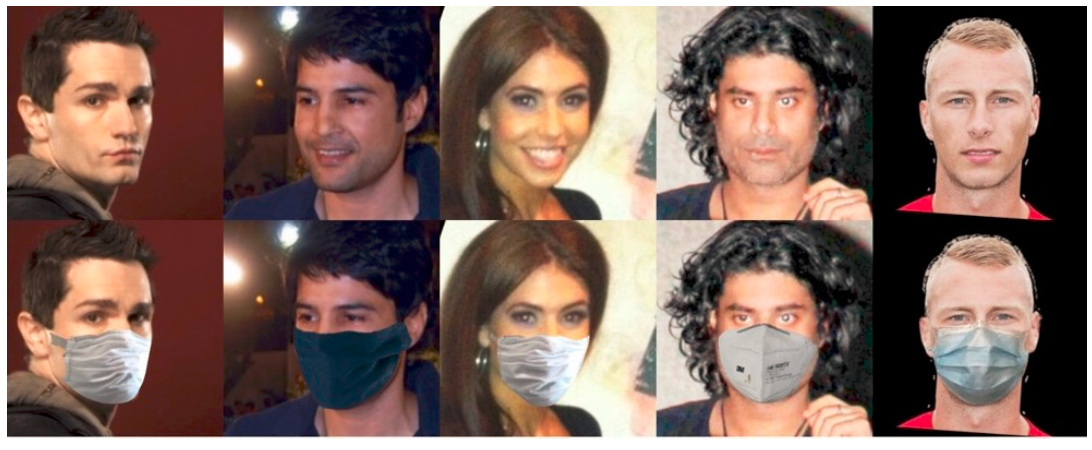}
    \caption{Example of synthetically generated masked face images.}
    \label{fig1}
\end{figure}

In stead of using public available MFR dataset, we use a synthetic tool to generate MFR dataset. The tool we use is the FMA-3D \cite{FMA-3D}. This tool can generate masked face images online or offline. We find that generating masked face images online affects the throughout of the training substantially, we so decide to generate masked face images offline. Shown in Figure \ref{fig1} are some examples of the generated masked face images.  

\subsection{Network Architecture}

\begin{figure*}[t]
    \centering
    \includegraphics[width=18cm]{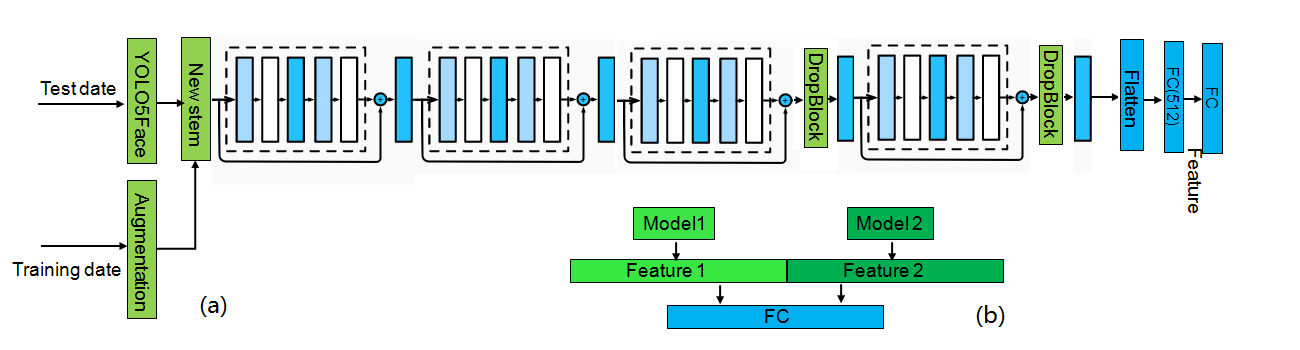}
    \caption{Network architecture. All key changes are highlighted in green. }
    \label{network}
\end{figure*}

We use the ArcFace framework \cite{ArcFace} with Resnet \cite{Resnet} as backbone. The network architecture is shown in Figure \ref{network}. All changed blocks are highlighted in green. Figure 2 (a) is the overall architecture, where YOLO5Face \cite{YOLO5Face} is used only on the test data. The training data provided by Webface \cite{webface260m} or Glint360k \cite{glint360} are used with data augmentation. The final feature map is flatten, then a 512-neuron full-connection (FC) layer is used to generate the feature. Lastly another FC layer is used for classification. Figure 2 (b) is the feature concatenation of two models, which will be described later.       

\subsection{Stem Unit}

In face recognition backbone likes ResNet \cite{Resnet} there is a stem unit - a component whose goal is to quickly reduce the input resolution. Typically, the input is transformed from [b, c, h, w] to [b, c*2, h//2, w//2], where b,c,h,w are batch,channel,height,and width of the input images. ResNet50 \cite{Resnet} stem is comprised of a stride-2 conv7x7 followed by a max pooling layer which reduces the input resolution by a factor of 4. The ResNet50-D \cite{ResNet-D} stem design is more elaborate - the conv7x7 is replaced by three conv3x3 layers. The new ResNet50-D stem design improves accuracy, but at a cost of lowering the training throughput. In the TResNet \cite{TResNet}, the stem unit is called a DepthToSpace transformation layer, which aims to replace the traditional convolution-based downscaling unit by a fast and seamless layer, with little information loss as possible. In YOLOv5 \cite{YOLOv5} the author wants to reduce the cost of Conv2d computation and use a tensor reshaping to reduce space resolution and increase the depth. It has been shown that this focus layer has a good impact on the YOLOv5 performance. 

\begin{figure}[t]
    \centering
    \includegraphics[scale=0.7]{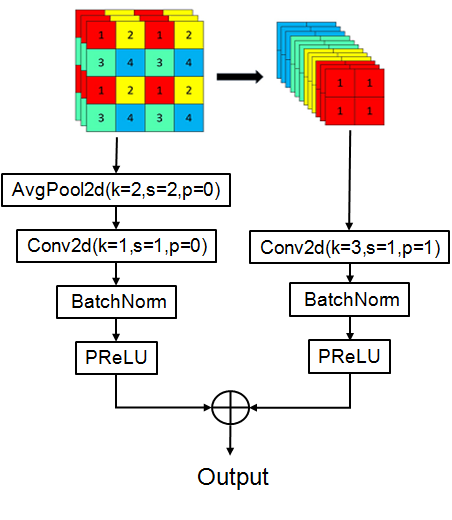}
    \caption{Architecture of the stem unit. The parameters are kernel (k), stride (s), and padding (p). }
    \label{fig2}
\end{figure}

Inspired by the stem unit in Resnet50 \cite{Resnet} and the focus layer in the YOLOv5 \cite{YOLOv5}, we design a new stem unit of down-sampling rate 2 similar to the focus layer in YOLOv5. There are two parallel stem units C1 and C2, whose output are added up. In the first unit C1, the input image is first passed to an average-pooling layer with kernel size=2 and stride=2 to reduce the space resolution. After that, a Conv layer with kernel size=2, stride=2 is applied. In the second unit C2, the input image is 2x down-sampled on the dimensions h and w. The even and odd index on the dimension h and w are mutually combined to form four images. These four images are concatenated to form an image whose resolution is 2x reduced but the depth is 4x increased. This image is sent to a Conv layer whose kernel=3x3, stride=2, padding=1.  In both stem units, a batchNorm layer and a PReLU layer are used after the Conv layer.  The architecture of this stem unit is shown in Figure \ref{fig2}.   

\subsection{DropBlock}

In deep learning, dropout is a widely used regularization method. However, it is often less effective for convolutional layers. This is perhaps due to the fact that activation units in convolutional layers are spatially correlated so information can still flow through convolutional networks despite dropout. Thus a structured form of dropout is needed to regularize convolutional networks. In \cite{dropblock}, A DropBlock, where units in a contiguous region of a feature map are dropped, is proposed. It is found that applying the DropBlock increases the accuracy. 

We add this DropBlock into our face recognition network. This block is added after the Conv layer and the skip connection in the last two layers of the Resnet \cite{Resnet}, as shown in Figure 1 (a).   

\subsection{Face Alignment using YOLO5Face} 

Face images are aligned using landmarks before they are sent to face recognition. Before RetinaFace becomes available, MTCNN \cite{MTCNN} is widely used for face image alignment. In Webface dataset, the RetinaFace \cite{RetinaFace} is used as the standard for face alignment. YOLO5Face \cite{YOLO5Face} is a delicate redesigned face detector from the YOLOv5 \cite{YOLOv5} object detector. In addition to the bounding box and confidence score, it also outputs five-point landmark, similar to MTCNN and RetinaFace. We use the YOLO5Face for face image alignment in some of our studies. 

\begin{figure}[t]
    \centering
    \includegraphics[scale=0.23]{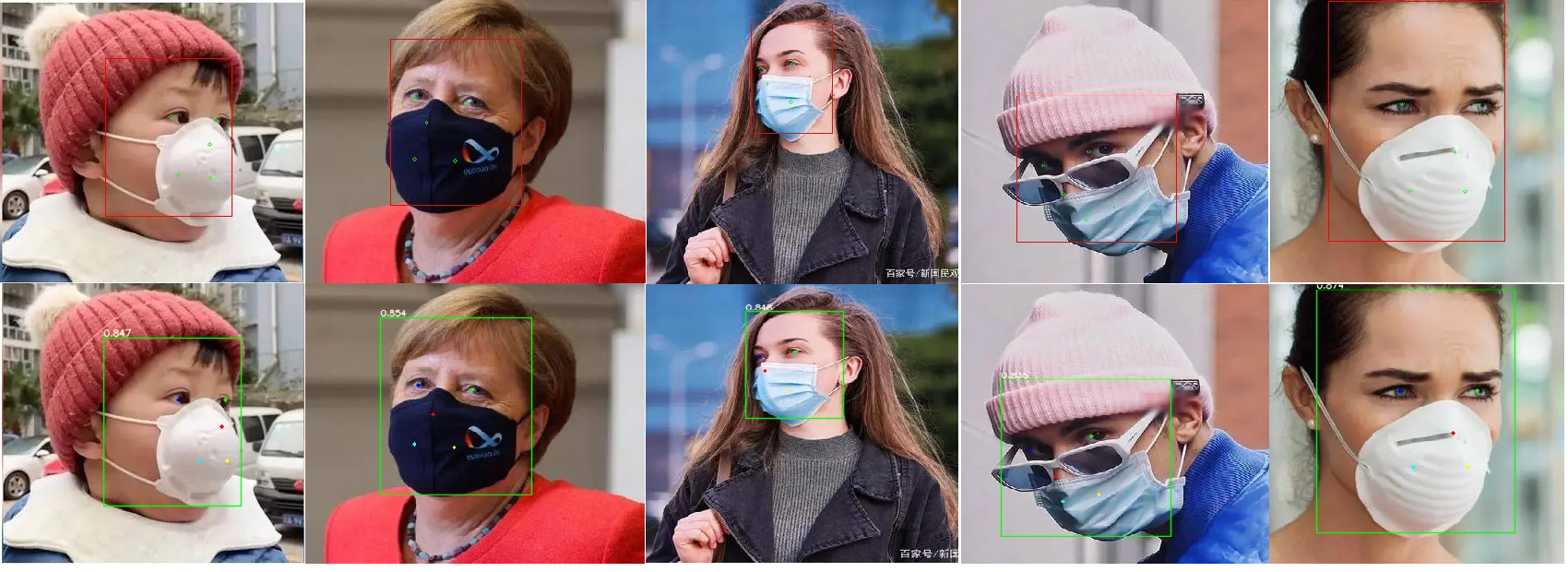}
    \caption{Examples of landmark, first row is from RetinaFace \cite{RetinaFace}, and second row from the YOLO5Face \cite{YOLO5Face}.}
    \label{fig3}
\end{figure}

Some qualitative examples are shown in Figure \ref{fig3}. The landmarks from the YOLO5Face \cite{YOLO5Face} and RetinaFace \cite{RetinaFace} on a set of face images with large pose are shown. It is not hard to see that the landmarks of YOLO5Face are better. Please note that the some of the detection scores are small because these faces have large poses.   

\subsection{Feature Concatenation} 
The left-right face flipping has been shown to improve the face recognition performance. It is not allowed to use in this MFR competition. We borrow this idea to use extracted feature vectors from multiple models. We choose two best models, extract their feature vectors, and concatenate them as the feature vector for the face recognition, as shown in Figure 2 (b). This is like an model ensemble concept. Our experiment shows that the performance can be improved.  

\section{Experiments}

\subsection{Dataset}

In early time of the first phase, the Glint360K dataset \cite{glint360} is used as the preliminary training dataset. After we meet the required baseline performance, 30\% of the WebFace260M \cite{webface260m} are released however are not used in our training. Other than the masked face images already included in the dataset, more masked face images are generated, as described in Section 3.5. The total number of masked face images is not exceeding 10\% of the total number of face images. 

\subsection{Implementation details}

The ArcFace framework \cite{ArcFace} in Pytorch is used in our study. Input image size is set to 112x112. We use the Resnet34 \cite{Resnet} as our backbone. We expect larger models like Resnet50, Resnet101 \cite{Resnet} will give better performance. But for faster training speed, we use Resnet34 in most of our studies.   

Other data augmentation methods we use include random cropping, random flipping, and the more complex Albumentations \cite{Albumentations}, which include Affine transforms (scaling, translation, rotation, distortion), noise, blurring, brightness and contrast adjustment.   

\begin{figure}[t]
    \centering
    \includegraphics[scale=0.5]{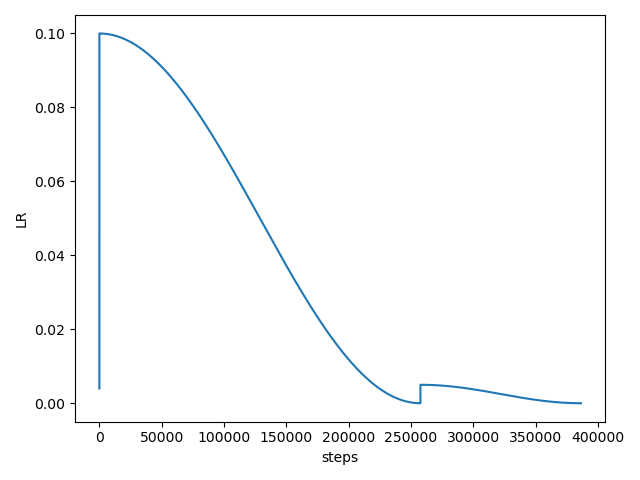}
    \caption{Cycle cosine decay learning rate, where one epoch is around 16100 steps.}
    \label{fig4}
\end{figure}

Our model is trained on four Nvidia RTX GPUs. The training runs 24 epochs with batch size 512. We use the cyclic cosine decay \cite{sgdr, consistent} for the learning rate strategy as shown in Figure \ref{fig4}. The initial learning rate is 0.1, the momentum is 0.9, and the weight decay is 5E-4. We use 0.1 epoch as warm-up, the learning rate is reduced to the minimum decay learning rate in 16 epochs.  

\subsection{Ablation Study}

In the Insightface/Glint360k track, we use Resnet34 \cite{Resnet} backbone for its fast training speed. We use this track as our ablation study tool in addition to submission to the competition leader board. 

\begin{table}[tb]
    \centering
    \begin{tabular}{c|c|c|c}
        \hline
        Technique  & MFR-Mask & MFR-All & Time\\
        \hline
        Baseline & 65.10 & 83.02 & 3.285 \\
        + Cycle Cosine LR & 67.38 & 83.08 & 3.285 \\
        + Data Aug & 77.20 & 82.65 & 3.285 \\
        + Stem Unit & 77.13 & 82.45 & 3.165 \\
        + SE & 78.37 & 83.12 & 4.294 \\
        + DropBlock & 79.21 & 84.22 & 4.294 \\
        + EMA & 79.35 & 84.64 & 4.294\\
        \hline
    \end{tabular}
    \caption{Ablation study on the Insightface/Glint360k track with backbone Resnet34 \cite{Resnet}. Time refers to the average inference time of the model.}
    \label{t1}
\end{table}

We test a variety of techniques in this study in an incremental manner, and the results are listed in Table 1. In the table, the cyclic cosine learning rate is described in Section 4.2; data augmentation refers to random cropping plus the Albumentations augmentation with random horizontal flipping, blurring, Gaussian blurring, motion blurring, RGB shift, and image compression, where the probability is 0.5 for the horizontal flipping, and 0.05 for all others. We also test the Squeeze-and-Excitation (SE) network \cite{SE}, and the Exponential Moving Average (EMA) gradient update.   

We see that the cycle cosine learning rate brings a 2.2\% performance improvement; the data augmentation brings the biggest improvement of nearly 10\%; all other techniques brings 0.1-0.2\% improvement except for the stem unit. The stem unit does not improve the accuracy, but it improves the inference time.   

\textbf{Face alignment with YOLO5Face \cite{YOLO5Face}.} We do this ablation study on the Webface track. We use the best configuration from the previous ablation study as baseline. Instead of Resnet34, Resnet124 is used as backbone. The results are listed in Table 2. First we train two models $R124\_1$ and $R124\_2$, then we concatenate the features from them and form the third model. In these three  models, we use the RetinaFace  \cite{RetinaFace} face detector and alignment. Then we keep using the concatenation model, and replace the RetinaFace with the YOLO5Face \cite{YOLO5Face}. We test two models, a small-sized model YOLO5-S, and a medium sized model YOLO5-M. Both models give better performance than the third model. This demonstrates that the YOLO5Face gives better face detection and landmark prediction as we qualitatively show in Figure 3.    

\begin{table}[tb]
    \centering
    \begin{tabular}{c|c|c|c}
        \hline
        Model & FaceDetect & All-Masked & Dim \\
        \hline
        R124\_1 & Retina & 0.1132 & 512 \\
        R124\_2 & Retina & 0.1142 & 512 \\
        $R124\_1\|R124\_2$ & Retina & 0.1065 & 1024 \\
        $R124\_1\|R124\_2$ & YOLO5-S & 0.1061 & 1024 \\
        $R124\_1\|R124\_2$ & YOLO5-M & 0.1056 & 1024 \\
        \hline
    \end{tabular}
    \caption{Ablation study on the Webface track with backbone Resnet124 \cite{Resnet}. Retina refers to the RetinaFace \cite{RefineFace}. Two YOLO5Face models \cite{YOLO5Face}, a small model YOLO5-S, and a medium model YOLO5-M are used. The symbol $\|$ denotes feature concatenation.}
    \label{t2}
\end{table}

\subsection{Benchmark Results}

To avoid overfitting on masked/standard face recognition, the Webface track organizer decide to revise the formula for calculating all three MFR metrics in the leaderboard. New MFR metrics are designed to show a weighted sum to consider both masked and standard faces at the same time. The new formulas are shown below, 

\begin{itemize}
\item{New All-Masked (MFR) = 0.25 * Old All-Masked (MFR) + 0.75 * All (SFR).}
\item{New Wild-Masked (MFR) = 0.25 * Old Wild-Masked (MFR) + 0.75 * Wild (SFR).}
\item{New Controlled-Masked (MFR) = 0.25 * Old Controlled-Masked (MFR) + 0.75 * Controlled (SFR).}
\end{itemize}

As of August 3, 2021, our best results on the two tracks are listed in Table 3. 

\begin{table}[tb]
    \centering
    \begin{tabular}{c|c|c}
        \hline
        Track & Metric &  Time \\
        \hline
        Insightface/Glint260k & (84.327, 92.702)  & 17.835 \\
        Webface & (0.1056, 0.0187) & 996.00 \\
        \hline
    \end{tabular}
    \caption{Benchmark results. For the Insight/Glint360 track, the metrics are (Mask, MR-All), and the time is the inference time in millisecond. For the Webface track, the metrics are (All-Masked MFR, All SFR), and the time is total time in millisecond. }
    \label{t3}
\end{table}

\section{Conclusions}

In this paper we present a solution for the ICCV2021 MFR challenge \cite{MFR}. Our solution has a good balance of the masked MFR and the standard face recognition.     

{\small
\bibliographystyle{ieee_fullname}
\bibliography{iccvcovid}

\begin{thebibliography}{10}\itemsep=-1pt

\bibitem{glint360}
Xiang An, Xuhan Zhu, Yang Xiao, Lan Wu, Ming Zhang, Yuan Gao, Bin Qin, Debing
  Zhang, and Ying Fu.
\newblock Partial fc: Training 10 million identities on a single machine.
\newblock {\em arXiv, 2010.05222}, 2021.

\bibitem{anwar2020masked}
Aqeel Anwar and Arijit Raychowdhury.
\newblock Masked face recognition for secure authentication.
\newblock {\em arXiv,2008.11104}, 2020.

\bibitem{singlecameraMFR}
Vivek Aswal, Omkar Tupe, Shifa Shaikh, and Nadir~N. Charniya.
\newblock Single camera masked face identification.
\newblock In {\em 2020 19th IEEE International Conference on Machine Learning
  and Applications (ICMLA)}, pages 57--60, 2020.

\bibitem{consistent}
Ben Athiwaratkun, Marc Finzi, Pavel Izmailov, and Andrew~Gordon Wilson.
\newblock There are many consistent explanations of unlabeled data: Why you
  should average.
\newblock {\em arXiv,1806.05594}, 2019.

\bibitem{Albumentations}
Alexander Buslaev, Vladimir~I. Iglovikov, Eugene Khvedchenya, Alex Parinov,
  Mikhail Druzhinin, and Alexandr~A. Kalinin.
\newblock Albumentations: Fast and flexible image augmentations.
\newblock {\em Information}, 11(2), 2020.

\bibitem{vggface2}
Qiong Cao, Li Shen, Weidi Xie, Omkar~M. Parkhi, and Andrew Zisserman.
\newblock Vggface2: A dataset for recognising faces across pose and age.
\newblock {\em arXiv,1710.08092}, 2018.

\bibitem{deng2021mfrinsightface}
Jiankang Deng, Jia Guo, Xiang An, Zheng Zhu, and Stefanos Zafeiriou.
\newblock Masked face recognition challenge: The webface260m track report.
\newblock {\em ICCV Workshop}, 2021.

\bibitem{ArcFace}
Jiankang Deng, Jia Guo, Niannan Xue, and Stefanos Zafeiriou.
\newblock Arcface: Additive angular margin loss for deep face recognition.
\newblock {\em CVPR}, June 2019.

\bibitem{RetinaFace}
J. Deng, J. Guo, Y. Zhou, J. Yu, I. Kotsia, and S. Zafeiriou.
\newblock Retinaface: Single-stage dense face localisation in the wild.
\newblock {\em CVPR}, 2020.

\bibitem{LPD}
Feifei Ding, Peixi Peng, Yangru Huang, Mengyue Geng, and Yonghong Tian.
\newblock Masked face recognition with latent part detection.
\newblock {\em ACM MM}, 2020.

\bibitem{FMA-3D}
FMA-3D.
\newblock Fma-3d.
\newblock {\em
  github.com/JDAI-CV/FaceX-Zoo/tree/main/addition\_module/face\_mask\_adding/FMA-3D},
  2021.

\bibitem{dropblock}
Golnaz Ghiasi, Tsung-Yi Lin, and Quoc~V. Le.
\newblock Dropblock: A regularization method for convolutional networks.
\newblock {\em arXiv,1810.12890}, 2018.

\bibitem{hariri2021efficient}
Walid Hariri.
\newblock Efficient masked face recognition method during the covid-19
  pandemic.
\newblock {\em arXiv,2105.03026}, 2021.

\bibitem{Resnet}
K. He, X. Zhang, S. Ren, and J. Sun.
\newblock Deep residual learning for image recognition.
\newblock {\em CVPR}, 2016.

\bibitem{ResNet-D}
T. He, Z. Zhang, H. Zhang, Z. Zhang, J. Xie, and M. Li.
\newblock Bag of tricks for image classification with convolutional neural
  networks.
\newblock {\em CVPR}, 2019.

\bibitem{SE}
Jie Hu, Li Shen, Samuel Albanie, Gang Sun, and Enhua Wu.
\newblock Squeeze-and-excitation networks.
\newblock {\em arXiv,1709.01507}, 2019.

\bibitem{MFR}
ICCV21-MFR.
\newblock Iccv21-mfr workshop.
\newblock {\em https://www.face-benchmark.org/challenge.html}, 2021.

\bibitem{cropping_attention}
Y. Li, K. Guo, Y. Lu, and et al.
\newblock Cropping and attention based approach for masked face recognition.
\newblock {\em Appl Intell}, 2021.

\bibitem{eqface}
R. Liu and W. Tan.
\newblock Eqface: A simple explicit quality network for face recognition.
\newblock {\em CVPR Workshop}, 2021.

\bibitem{sgdr}
Ilya Loshchilov and Frank Hutter.
\newblock Sgdr: Stochastic gradient descent with warm restarts.
\newblock {\em arXiv,1608.03983}, 2017.

\bibitem{montero2021boosting}
David Montero, Marcos Nieto, Peter Leskovsky, and Naiara Aginako.
\newblock Boosting masked face recognition with multi-task arcface.
\newblock {\em arXiv,2104.09874}, 2021.

\bibitem{YOLO5Face}
D. Qi, W. Tan, Q. Yao, and J. Liu.
\newblock Yolo5face: Why reinventing a face detector.
\newblock {\em ArXiv,2105.12931}, 2021.

\bibitem{TResNet}
T. Ridnik, H. Lawen, A. Noy, E.~B. Baruch, G. Sharir, and I. Friedman.
\newblock Tresnet: High performance gpu-dedicated architecture.
\newblock {\em WACV}, 2021.

\bibitem{universal}
Yichun Shi, Xiang Yu, Kihyuk Sohn, Manmohan Chandraker, and Anil~K. Jain.
\newblock Towards universal representation learning for deep face recognition.
\newblock {\em CVPR}, 2020.

\bibitem{COVID-trace}
W. Tan and J. Liu.
\newblock Application of face recognition in tracing covid-19 patients and
  close contacts.
\newblock {\em IEEE ICMLA}, 2020.

\bibitem{CosFace}
Hao Wang, Yitong Wang, Zheng Zhou, Xing Ji, Dihong Gong, Jingchao Zhou, Zhifeng
  Li, and Wei Liu.
\newblock Cosface: Large margin cosine loss for deep face recognition.
\newblock {\em CVPR}, 2018.

\bibitem{survey}
Mei Wang and Weihong Deng.
\newblock Deep face recognition: A survey.
\newblock {\em Neurocomputing}, 429:215--244, 2021.

\bibitem{NAN}
Y. Yang, P Ren, D. Zhang, D. Chen, F. Wen, H. Li, and G. Hua.
\newblock Neural aggregation network for video face recognition.
\newblock {\em CVPR}, 2017.

\bibitem{YOLOv5}
YOLOv5.
\newblock Yolov5.
\newblock {\em https://github.com/ultralytics/yolov5}, 2020.

\bibitem{MTCNN}
K. Zhang, Z. Zhang, Z. Li, and Y. Qiao.
\newblock Joint face detection and alignment using multitask cascaded
  convolutional networks.
\newblock {\em IEEE Signal Processing Letters}, 23(10):1499--1503, 2016.

\bibitem{RefineFace}
S. Zhang, C. Chi, Z. Lei, and S.Z. Li.
\newblock Refineface: Refinement neural network for high performance face
  detection.
\newblock {\em ArXiv preprint 1909.04376}, 2019.

\bibitem{zhu2021mfrwebface}
Zheng Zhu, Guan Huang, Jiankang Deng, Yun Ye, Junjie Huang, Xinze Chen, Jiagang
  Zhu, Tian Yang, Jia Guo, Jiwen Lu, Dalong Du, and Jie Zhou.
\newblock Masked face recognition challenge: The webface260m track report.
\newblock 2021.

\bibitem{webface260m}
Zheng Zhu, Guan Huang, Jiankang Deng, Yun Ye, Junjie Huang, Xinze Chen, Jiagang
  Zhu, Tian Yang, Jiwen Lu, Dalong Du, and Jie Zhou.
\newblock Webface260m: A benchmark unveiling the power of million-scale deep
  face recognition.
\newblock {\em CVPR}, 2021.

\end{thebibliography}
}

\end{document}